\documentclass[runningheads]{llncs}
\usepackage{graphicx}
\usepackage{amssymb}
\usepackage{amsmath}
\usepackage{textcomp}
\usepackage{booktabs, multirow}
\usepackage{hyperref}
\usepackage{algorithmic}

\usepackage{todonotes}
\usepackage{mathtools}
\DeclareMathOperator*{\argmin}{argmin} 

\begin{document}
\title{DeepRecon: Joint 2D Cardiac Segmentation and 3D Volume Reconstruction via A Structure-Specific Generative Method}

%
\titlerunning{DeepRecon}
%
\author{Qi Chang\inst{1} \and
Zhennan Yan\inst{2} \and
Mu Zhou\inst{2} \and Di Liu\inst{1} \and Khalid Sawalha\inst{3} \and Meng Ye\inst{1} \and Qilong Zhangli\inst{1} \and Mikael Kanski\inst{4} \and Subhi Al'Aref\inst{3} \and Leon Axel\inst{5} \and Dimitris Metaxas\inst{1}}

\authorrunning{Q. Chang et al.}
%

\institute{Rutgers University, Piscataway, NJ 08854, USA
\email{qc58@rutgers.edu}\\
 \and
SenseBrain Research \\
\and 
Department of Medicine, Division of Cardiology. University of Arkansas for Medical Sciences, Little Rock, AR, USA \\
\and
Department of Clinical Physiology, Skane University Hospital Lund, Lund University, 221 85 Lund, Sweden \\
\and
Department of Radiology, New York University, New York, NY 10016, USA.
}

\maketitle              
\begin{abstract}
Joint 2D cardiac segmentation and 3D volume reconstruction are fundamental in building statistical cardiac anatomy models and understanding functional mechanisms from motion patterns.
However, due to the low through-plane resolution of cine MR and high inter-subject variance, accurately segmenting cardiac images and reconstructing the 3D volume are challenging.
In this study, we propose an end-to-end latent-space-based framework, DeepRecon, that generates multiple clinically essential outcomes, including accurate image segmentation, synthetic high-resolution 3D image, and 3D reconstructed volume. 
Our method identifies the optimal latent representation of the cine image that contains accurate semantic information for cardiac structures. 
In particular, our model jointly generates synthetic images with accurate semantic information and segmentation of the cardiac structures using the optimal latent representation.
We further explore downstream applications of 3D shape reconstruction and 4D motion pattern adaptation by the different latent-space manipulation strategies. 
The simultaneously generated high-resolution images present a high interpretable value to assess the cardiac shape and motion.
Experimental results demonstrate the effectiveness of our approach on multiple fronts including 2D segmentation, 3D reconstruction, downstream 4D motion pattern adaption performance.



\end{abstract}
\section{Introduction}
\label{sec:intro}
Comprehensive image-based assessment of cardiac structure and motion through 3D heart modeling is essential for early detection, cardiac function understanding, and treatment planning of cardiovascular diseases (CVD) \cite{awori20213d,kustner2020cinenet}. 
As a standard clinical diagnostic tool, cine magnetic resonance imaging (cMRI) has been used to characterize the complex shape and motion of the heart. 
cMRI presents multiple advances, including the high temporal and in-plane resolution, minimal radiation exposure, and improved soft tissue definition \cite{prakash2010multimodality}. 
Yet conventional practice acquires a stack of 2D short-axis (SAX) slices with large between-slice spacing, so the accurate heart tissue segmentation and 3D image-based heart modeling is challenging due to the missing structure information.


Heart modeling from cMRI typically consists of carefully-designed steps, including image segmentation \cite{bernard2018deep,liu2021refined,chang2020soft} and 3D reconstruction~\cite{biffi20193d}. 
Deep learning has shown its progress in the segmentation of cardiac structures \cite{bernard2018deep,campello2021multi}, which addresses the challenges of analyzing the complex and variable shape of the heart and ill-defined borders in MR images. 
After obtaining 2D segmentation, 3D reconstruction can be implemented by subsequent interpretations between neighboring slices. Conventional approaches \cite{frakes2008new,leng2013medical} often struggle when significant anatomical changes appear in the consecutive slices. 
For instance, deformable models have been proposed for 3D surface construction of left ventricular (LV) wall and motion tracking \cite{yu2014deformable,ye2021deeptag}. Yet the parameter initialization is sensitive, making it difficult to generalize across clinical settings. Despite that 3D high-resolution image acquisition and generation are gaining momentum to assess disease status~\cite{biffi20193d,kustner2020cinenet,xia2021super}, the integrative analysis of MR data enables high-quality MR image segmentation, reconstruction, and subsequent interpretation has not been explicitly addressed.

In this study, we propose an end-to-end, latent-space-based framework, DeepRecon, that generates multiple clinically essential outcomes, including accurate image segmentation, synthetic high-resolution 3D image, and 3D reconstructed volume (see Fig. 1). 
Our method could jointly generate 2D segmentation and 3D volume (by interpolating latent codes) simultaneously in the evaluation stage. Thus, we do not require another step to reconstruct the 3D volume from the sparse 2D segmentations. Meanwhile, the simultaneously generated synthetic images present a high interpretable value to assess the shape and motion of cardiology. Our study draws inspiration from StyleGANs \cite{karras2019style,karras2020analyzing}, where a synthetic image can be generated from a random latent code. 
Our findings are built upon the rationale that the latent code can be used to reconstruct the realistic synthetic image and simultaneously generate accurate segmentation~\cite{li2021semantic,richardson2021encoding}. Experimental results demonstrate the effectiveness of our approach on 2D segmentation, 3D reconstruction, and 4D motion pattern adaptation performance.






\section{Method}
\label{sec:method}
Fig.\ref{fig:arch} illustrates the end-to-end framework of DeepRecon that learns a latent space from MR imaging to yield a broad range of outcomes, including generation of high-quality 2D cine image and corresponding segmentation, high-resolution 3D reconstructed volume, and 4D motion adaptation of the heart. We first describe the architecture design and the learning process and then elaborate on latent-space-based 3D reconstruction and motion adaptation. 

\subsection{DeepRecon architecture}

\begin{figure}[t]
    \centering
    \includegraphics[width=1\textwidth]{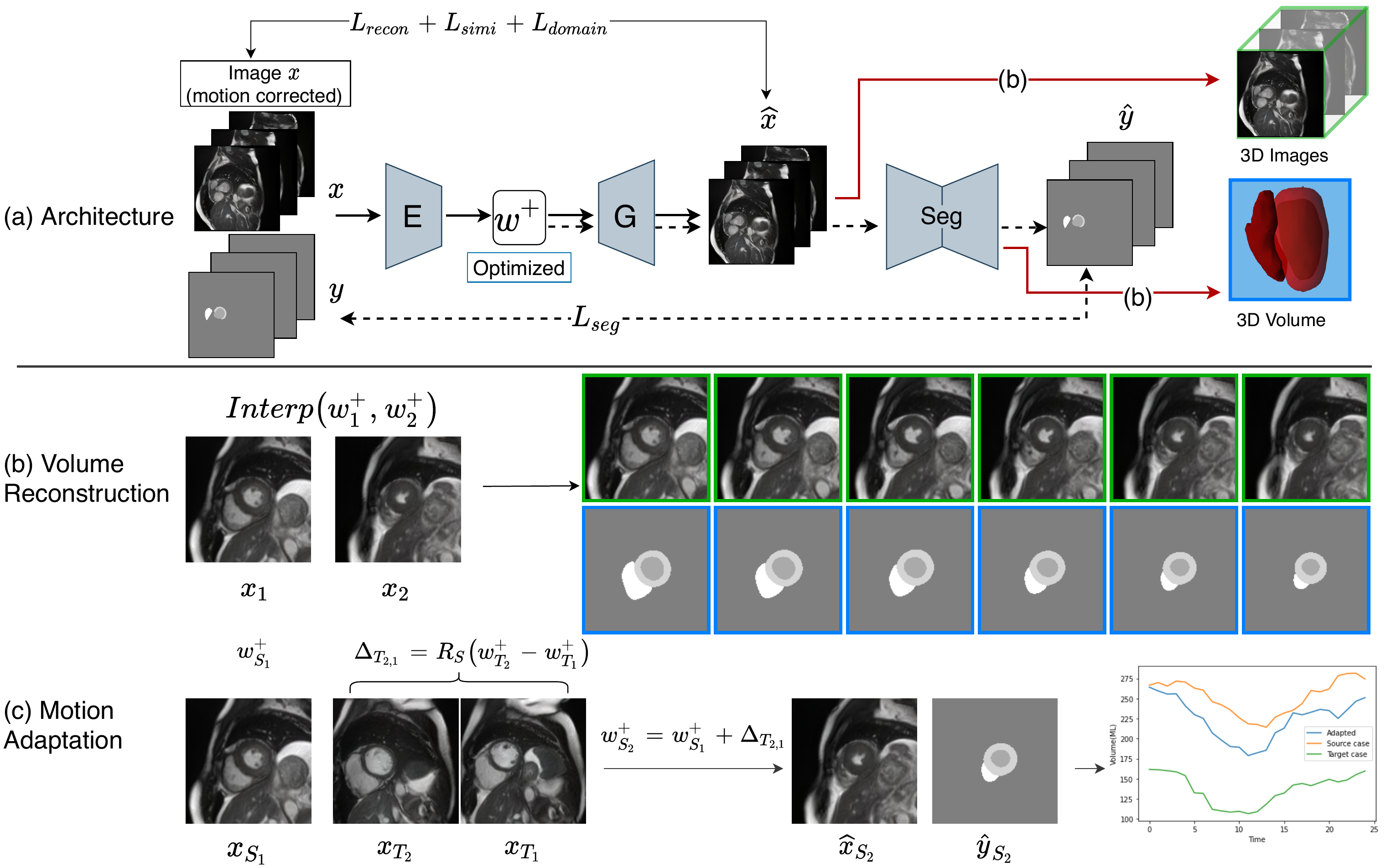}
    \caption{(a) illustrates the architecture of DeepRecon. The black arrow workflow demonstrates the hybrid inversion method to acquire the optimized latent code. The black dash workflow shows the segmentation training stage, and the red workflow shows the 3D image and volume reconstruction process. (b) shows a 3D volume reconstruction (both images and masks) by interpolating the latent code of adjacent slices $x_1$ and $x_2$. (c) shows a source case $x_{S_1}$ can adapt the motion pattern extracted from temporal frames $x_{T_2}$ and $x_{T_1}$ of a target case, so the source's synthetic motion has similar volume changes as the target. Best viewed in color.
    }
    \label{fig:arch}
\end{figure}

Our architecture consists of a latent-code encoder $E$, a MR image generator $G$ followed by a segmentation network $Seg$, as shown in Fig.\ref{fig:arch} (a). The $E$ takes an real MR image $x$ as input and outputs its latent code representation $w^+$. The optional motion correction is described in \cite{yang20173d,chang2021unsupervised}. Then, $G$ can generate synthetic image $\hat x$ from the $w^+$, and $Seg$ produces the segmentation $\hat y$. 
As an intuitive downstream application, we can interpolate the latent codes of two neighboring MR slices to produce 3D images and masks by stacking continuous synthetic outputs (spatial domain synthesize), as shown in Fig.\ref{fig:arch} 
(b). Furthermore, we show that the manipulation of latent codes enables a 4D motion adaptation that can transfer a motion pattern of one subject to another subject (temporal domain synthesize), 
as shown in Fig. \ref{fig:arch} (c) and described in section \ref{sec:3drecon}. 

\subsection{Learning of DeepRecon}
The training process of DeepRecon includes three stages: training of the generator $G$ and the encoder $E$, hybrid optimization for the latent code, and latent space based segmentation module. 
The training of the $G$ and $E$ follow the methods in \cite{karras2020analyzing} and \cite{richardson2021encoding}, respectively.
Then, we use GAN inversion to find the optimal latent codes for real cine SAX images. We adopt a hybrid inversion strategy for efficiency and accuracy. Given a target image $x$, we predict the latent code $w^+=E(x)$. The $w^+$ is then optimized by reconstructing the target image.
We formulate the objective function of the optimization by three losses, including a reconstruction term $L_{smooth}$, a similarity term $L_{simi}$, and a domain regularization term $L_{domain}$.
\begin{equation}
    \label{equ:opt}
    w^{+*} = \argmin\limits_{w^+}\mathcal L(x,G(w^+)) \\
    = \argmin\limits_{w^+} (\lambda_1 \mathcal L_{recon} +\lambda_2 \mathcal L_{simi} + \lambda_3 \mathcal L_{domain})
\end{equation}
\begin{equation}
    \label{equ:loss}
    \begin{aligned}
    \mathcal L_{recon} =& \mathcal L_{LPIPS}(x, G(w^+)) +
    \lVert ROI(x-G(w^+)) \rVert_2^2, \\
    \mathcal L_{simi} =& -L_{NCC}(x, G(w^+)), 
    \mathcal L_{domain}=\lVert w^+ - E(G(w^+))\rVert^2_2
    \end{aligned}
\end{equation}
where $\lambda_1, \lambda_2, \lambda_3$ are balancing hyperparameters.
For the reconstruction term, we use the Learned Perceptual Image Patch Similarity (LPIPS) distance \cite{zhang2018unreasonable} to encourage smooth generator mapping from latent codes. We also include a ROI-based regularization by a weighted L2 term to encourage the similarity in the ROI. 
For the similarity term, we adopt a normalized local cross-correlation (NCC) metric\cite{avants2011reproducible} to ensure robustness for intensity-variant cine images.
The last term regularizes the optimization trajectory to stay in the training domain and keep the interpolation of latent-space vectors smoothness \cite{zhu2020domain}. 


We store the optimized latent codes of training samples for training the $Seg$ (e.g., U-Net \cite{ronneberger2015u}). The $Seg$ takes synthetic images $\hat x=G(w^{+*})$ as input and train the network with the Cross-Entropy(CE) and DICE loss\cite{sudre2017generalised,ronneberger2015u}. 
The loss of the segmentation task is defined as $\mathcal L_{seg} = \mathcal L_{ce} + \mathcal L_{dice}$. 

\subsection{3D reconstruction and 4D motion adaptation}
\label{sec:3drecon}
After learning of the DeepRecon, we use the learned model towards 3D shape reconstruction and 4D motion adaptation that take fully advantage of the learned latent space for the MR images.
Our key motivation is that, by identifying the optimal latent-space representation, the spatial interpolation of two neighboring SAX slices can be achieved by generating images from the interpolation of their corresponding latent codes (see Fig. \ref{fig:arch} (b)).
In this way, DeepRecon is able to output a smooth super-resolution cine image volume and the corresponding 3D heart shape from the generated segmentations.


Beyond the spatial interpolation, we can adapt a motion pattern of a target subject $x_T$ to a source subject $x_S$. In this way, we can synthesize and visualize a healthy-heart motion of a diseased subject and compare with its real motion pattern in an intuitive manner.
The motion adaptation is demonstrated in Fig. \ref{fig:arch} (c). 
Formally, we define the motion adaptation module as:
\begin{equation}
\label{equ:adaption}
    \hat{x}_{S_{i+1}} = G(w^+_{S_0}+ \sum^i_{n=0}\mathcal R_{S}\Delta_{T_{n+1,n}}) 
\end{equation}
where $\hat{x}_{S_{i+1}}$ is the generated source images in time $i+1$; $w^+_{S_0}$ is the latent code for $x_{S_0}$; $\Delta_{T_{n+1,n}}=(w^+_{T_{n+1}}-w^+_{T_{n}})$ and $w^+_{T_n}$ is the latent code for target subject in time $n$. Because the size of heart varies, $\mathcal R_S$ represents the resampling function that align the number of target slices to the number of source slices.

\section{Experiments}
\label{sec:experiment}
We evaluate multiple outputs of DeepRecon including latent-space-based 2D segmentation (\ref{sec:2dseg}), 3D volume reconstruction (\ref{sec:exp3drecon}), and motion adaptation (\ref{sec:motionadapt}).


\textbf{Dataset:} The UK Biobank dataset \cite{petersen2015uk} consists of SAX and LAX cine CMR images of normal subjects. Cardiac structures, LV cavity (LVC), LV myocardium (LVM), and right-ventricle cavity (RVC) were manually annotated on SAX images at the end-diastolic (ED) and end-systolic (ES) cardiac phases \cite{petersen2015uk}.
We use 6,846 cases containing 3,569,990 2D SAX MR images as a pre-training subset. We adopt the first 1,010 annotated subjects and split them into two sets of 810/200 for training and validation of DeepRecon. We select another 100 cases from the UK Biobank as the testing set that has no overlap with the pre-training, training and validation subsets. Besides, two cardiologists annotated and verified the LAX images of 50 testing cases to evaluate the 3D volume reconstruction task.
Finally, we validate the motion adaptation on a private dataset with diseased cases to assess the motion patterns among various cardiac functional diseases.

\subsection{Latent-space-based 2D segmentation}
\label{sec:2dseg}
\textbf{Settings}:
We first evaluate the generative segmentation method based on the latent representation of MR images. The purpose of this task is to ensure the generated 2D segmentation is accurate enough to reconstruct 3D cardiac shape precisely.
We pre-train the MR image generator $G$ on the pre-training subset based on StyleGAN2 method~\cite{karras2020analyzing}. The training took 14 days with 4 RTX 8000 GPUs and achieved Fid50k score of 18.09.
We perform the following experiments for comparison.
1) $DirectSeg$. We train a segmentation network with the same architecture as the $Seg$ in our method directly on the real images. This result shows the approximate upper bound of the segmentation task.
2) $3D$-$Unet$. In this method, we evaluate the 3D-Unet segmentation on the original low-resolution Cine data. Since there is no high-resolution 3D data for training the 3D-UNet, it can not be directly applied to the generated high-resolution 3D images to obtain the 3D segmentation volume.
3) $SemanticGAN$ \cite{li2021semantic}. In this method, the architecture of the generator is extended to output the image and segmentation at the same time (no additional segmentation network is required). 
4) $W^+SegNet$. A variant of our method keeps the same architecture except the $Seg$ network takes latent code as input to directly predict the segmentation.
5) $DeepRecon_{no}$. The proposed method without the hybrid latent code optimization. 
6) $DeepRecon_{1k}$, $DeepRecon_{10k}$. 
These two settings of the proposed method use the latent code generated by the encoder $E$, followed by 1k and 10k optimization steps, respectively. 
We train all the models in 2D and use 3D DICE and 95\% quantile of Hausdorff distance (HD95) for evaluation.

\begin{figure}[tb]
    \centering
    \includegraphics[width=\textwidth]{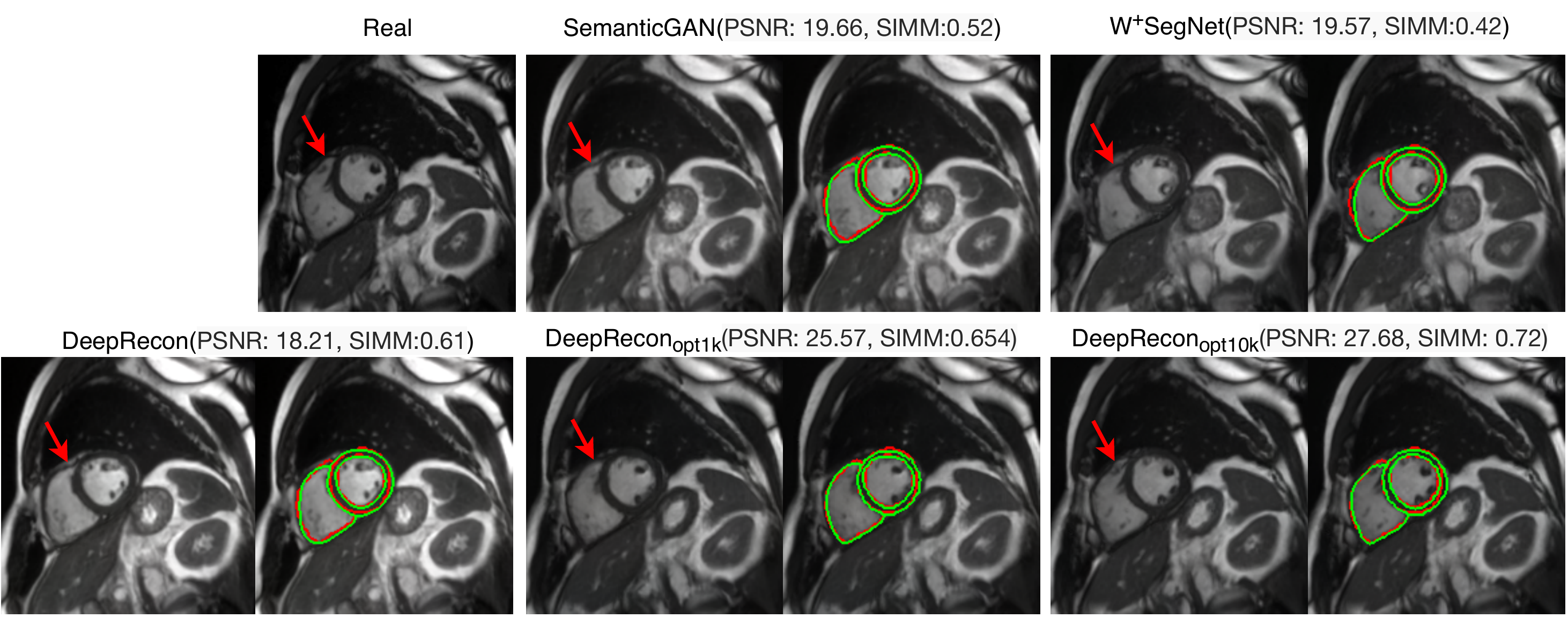}
    \caption{Segmentation results of different latent-space-based models. Each group shows the synthetic image and corresponding segmentation. The red lines are the ground truth boundaries, and the green lines are the predictions. We notice that the sub-optimal latent codes may differ in the cardiac region, especially thin structures. 
    }
    \label{fig:seg_sample}
\end{figure}

\textbf{Result}:
We report the quantitative segmentation results in Table \ref{tab:seg}, and illustrates a representative example in Fig. \ref{fig:seg_sample}. Table \ref{tab:seg} shows that DeepRecon consistently outperforms the other latent-space-based methods and achieves similar performance as the upper bound. 
The proposed $DeepRecon_{1k}$ performs the best overall, especially in HD95, because more accurate image reconstruction with fine details helps to optimize the segmentation network. The $DeepRecon_{10k}$ further improves the DICE score of LVM while it drops in terms of HD95 (may due to overfitting) and takes 10x time to retrieve the optimized latent code. Thus, the $DeepRecon_{1k}$ presents a good trade-off between the accuracy and efficiency and is used for evaluation in following experiments. 
We also observe that other latent-space-based methods tend to have significantly worse results for the LVM compared with the LVC and RVC. A possible reason is that the sub-optimal latent code predicted from the encoder is challenging to accurately reconstruct such a thin annular shape. 
Our $DeepRecon_{1k}$, $DeepRecon_{10k}$ are statistically better than SematicGAN in all metrics ($p<0.05$) and comparable with DirectSeg ($p>0.05$) in most metrics. 
Additionally, we measure the image quality of the ground-truth and synthetic images using the averaged peak signal-to-noise ratio (PSNR) and structural similarity index (SSIM) in Fig. \ref{fig:seg_sample}. The results show the improvement of the synthesized image quality. 

\begin{table}[t]
    \caption{Segmentation results by various latent-code-based methods compared with a direct segmentation from real images: mean (standard deviation).}
    \centering
    \begin{tabular}{c|c|c|c|c|c|c}
    \hline
        \multirow{2}{*}{Method} & 
        \multicolumn{2}{c|}{LVC} &
        \multicolumn{2}{c|}{LVM} &
        \multicolumn{2}{c}{RVC} \\
    \cline{2-7}
        & DICE$\uparrow$ & HD95$\downarrow$ & DICE$\uparrow$ & HD95$\downarrow$ & DICE$\uparrow$ & HD95$\downarrow$ \\
    \hline
        $DirectSeg$ & $0.928(0.04)$ & $2.31(9.8)$ & $0.861(0.03)$ & $1.38(3.5)$ & $0.892(0.04)$ & $4.69(15.8)$ \\
        $3D$-$UNet$ & $0.938(0.046)$ & $3.78(11.4)$ & $0.861(0.03)$ & $2.31(1.8)$ & $0.90(0.03)$ & $5.79(3.9)$ \\
        \hline
        $SemanticGAN$ & $0.891(0.05)$ & $5.23(7.1)$ & $0.787(0.04)$ & $3.11(1.9)$ & $0.843(0.05)$ & $4.77(4.2)$ \\
        $W^+SegNet$ &  $0.895(0.05)$ & $5.97(9.5)$ & $0.789(0.06)$ & $3.11(2.7)$& $0.822(0.06)$ & $6.33(6.2)$ \\
        $DeepRecon_{no}$ & $0.895(0.04)$ & $2.37(8.1)$& $0.792(0.04)$& $1.87(5.8)$ &$0.847(0.05)$ & $2.33(5.6)$\\
        $DeepRecon_{1k}$ & $\textbf{0.926(0.04)}$& $\textbf{1.25(0.8)}$ & $0.835(0.05)$& $\textbf{1.27(1.7)}$ & $0.883(0.04)$ & $\textbf{1.51(4.4)}$\\
        $DeepRecon_{10k}$ & $0.925(0.04)$& $3.15(14.8)$& $\textbf{0.858(0.03)}$& $2.14(9.1)$& $\textbf{0.890(0.04)}$ & $2.70(10.2)$\\
    \toprule
    \end{tabular}
    \label{tab:seg}
\end{table}

\subsection{3D volume reconstruction}
\label{sec:exp3drecon}
\textbf{Settings}: 
We evaluate the reconstructed 3D cardiac shape continuity from the interpolated latent codes. Since 3D MR image and segmentation volume are unavailable, we utilize the LAX annotations and calculate the overlapping accuracy (2D DICE) of the intersection areas, i.e., the intersection of the reconstructed 3D volume and the LAX slice. We correct the misalignment for each SAX slice based on the method described in \cite{yang20173d}. 
The LAX slices usually include 2-chamber (2ch), 3-chamber (3ch),and 4-chamber (4ch) views. The following experiments are performed: 1) Original. The intersections of tiles of the original SAX annotations and each LAX view. 2) Linear Interpolation. 3) Coherent Point Drifting (CPD). The CPD algorithm \cite{myronenko2010point} can construct 3D surface based on a 3D point cloud and the deformable registration. 
4) $DeepRecon_{no}$,$DeepRecon_{1k}$,$DeepRecon_{10k}$ experiments follow the same configuration described in \ref{sec:2dseg}.
The DICE scores on different views and their average score are computed.

\begin{figure}[t]
    \centering
    \includegraphics[width=1\textwidth]{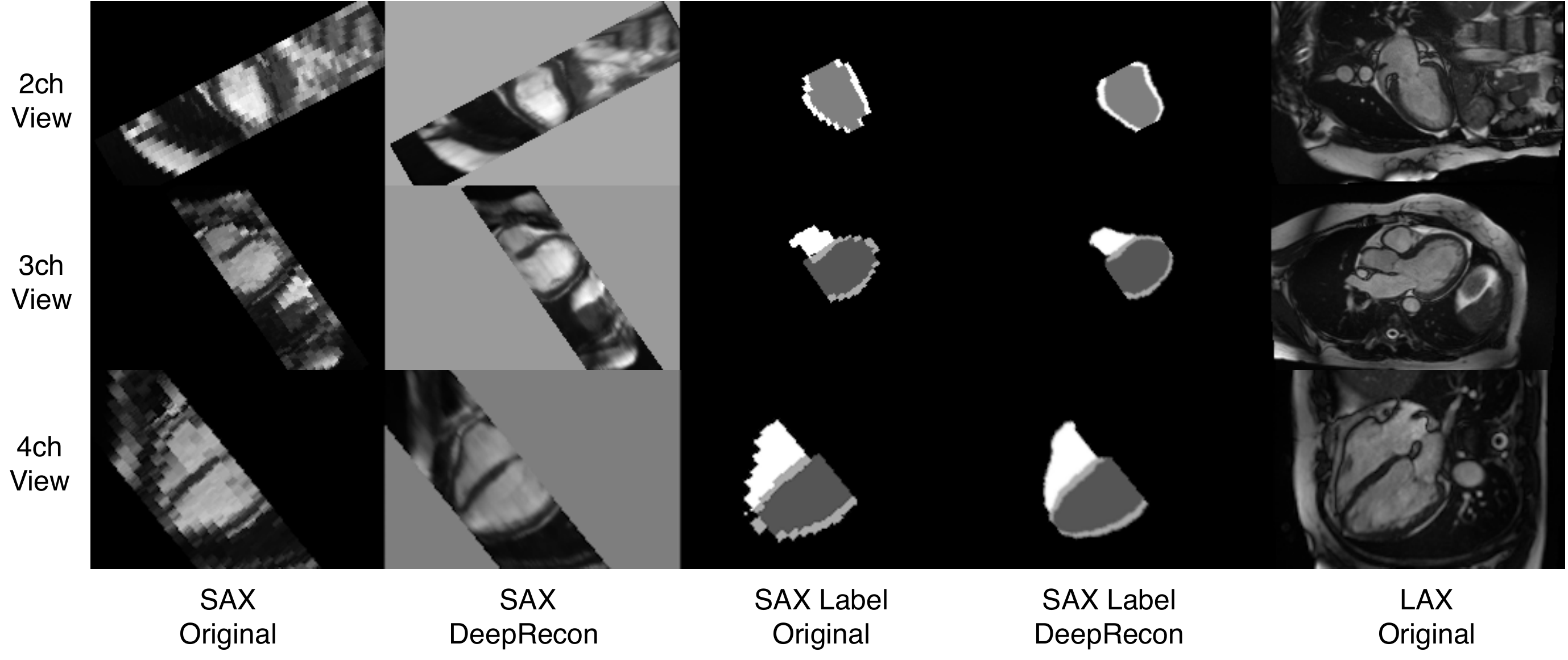}
    \caption{Visualization of the cross-sections of reconstructed 3D images and label volumes by DeepRecon in different LAX views (last column). The first and third columns show the original low-resolution SAX images and labels. The second and fourth columns show the high-resolution reconstructions from the proposed method.}
    \label{fig:recon3d}
\end{figure}

\textbf{Result:}
In Table \ref{tab:recon}, the DICE scores measure the accuracy for the LVC region between the LAX annotation and the intersection of 3D reconstructed volume and the corresponding LAX plane. We also show three examples in Fig. \ref{fig:recon3d} for each LAX view. Overall, our method achieves improved performance in each view compared with other approaches. 
We observe that the SAX slices are not complete compared to the LAX slices in the basal and apex region, but our methods still achieve a better performance in 2ch and 4ch views on average.

\begin{table}[t]
    \caption{Evaluation of reconstructed 3D volumes in terms of the DICE score of their intersections on each LAX plane.}
    \centering
    \begin{tabular}{c|c|c|c|c}
    \hline
        Method & Average DICE & 2ch view DICE & 3ch view DICE & 4ch view DICE\\
        \hline
        Original & $0.780\pm{0.111}$ & $0.787\pm{0.091}$ & $0.793\pm{0.105}$ & $0.766\pm{0.128}$\\
        Linear Interp& $0.781\pm{0.080}$ & $0.797\pm{0.051}$ & $0.773\pm{0.070}$ & $0.768\pm{0.102}$\\
        CPD\cite{myronenko2010point} & $0.790\pm{0.099}$&$0.803\pm{0.084}$ & $\textbf{0.815}\pm{\textbf{0.094}}$ & $0.767\pm{0.109}$ \\
    \hline
        $DeepRecon_{no}$& $0.806\pm{0.111}$ & $0.830\pm{0.069}$ & $0.799\pm{0.111}$ & $0.787\pm{0.091}$ \\ 
        $DeepRecon_{1k}$& $ \textbf{0.817}\pm{\textbf{0.097}}$ & $\textbf{0.848}\pm{\textbf{0.056}}$ & $0.802\pm{0.141}$ & $\textbf{0.797}\pm{\textbf{0.091}}$ \\ 
        $DeepRecon_{10k}$& $0.809\pm{0.094}$ & $0.839\pm{0.053}$ & $0.800\pm{0.136}$ & $0.787\pm{0.090}$ \\ 
    \toprule
    \end{tabular}
    \label{tab:recon}
\end{table}

\subsection{Motion pattern adaptation}
\label{sec:motionadapt}

\textbf{Settings:} We further perform an exploratory analysis to assess the cardiac 4D motion adaptation qualitatively. We extract a target case's motion pattern and assign such motion pattern to the initial shape of a source heart. First, we select two random healthy cases from the UK Biobank dataset. Second, we use one normal case as the target and one diseased case from our private dataset as the source. The diseased heart has shown severe mitral regurgitation with significant LV dilation and decreased function.

\textbf{Results:}
\begin{figure}[b]
    \centering
    \includegraphics[width=0.4\textwidth]{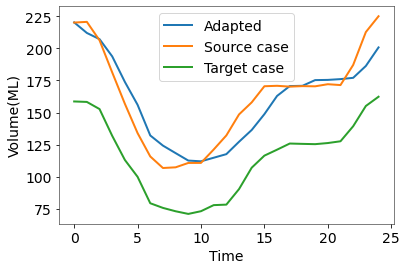}
    \includegraphics[width=0.4\textwidth]{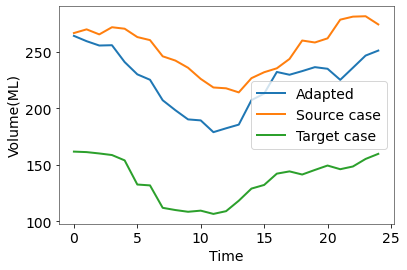}
    \caption{Two examples of motion adaptation: normal-to-normal (left) and diseased-to-normal (right). The motions are plotted by volume changes of LVC in a cardiac cycle.}
    \label{fig:motion}
\end{figure}

The volume changes of the LVC over time within a full cardiac cycle are presented in Fig. \ref{fig:motion}. The healthy example (Fig. \ref{fig:motion} left) shows the consistency of cavity volume change. By transferring the cardiac motion pattern from the target case to the source shape, the motion of the source heart (the blue Adapted curve) becomes similar to the target case (the green curve) while the heart scale remains the same.
In the second experiment, shown in Fig. \ref{fig:motion} right, we can synthesize a normal-like motion for the source diseased heart (the blue curve), which shows a more efficient contraction than its real motion (the orange curve). 
The ejection fraction of the source case is significantly smaller than the adapted motion. Such motion adaptation method provides a unique tool to potentially help cardiologists analyze functional differences between various cases (see video clips in the supplementary).

\section{Conclusion}
\label{sec:conclude}
Integrative analysis of cMRI is of great clinical significance in cardiac function understanding and assessment. This paper proposes an end-to-end latent-space-based generative method, DeepRecon, that generates multiple outcomes, including 2D image segmentation, 3D reconstructed volume, and extended motion pattern adaptation. Our findings show that the learned latent representation can lead to high-level performance on cMRI feature analysis. In particular, our approach opens up new perspectives on building scalable 3D/4D synthetic cardiac models for cardiac functional research. In the future, we will consider Transformer-based approaches \cite{gao2021utnet,liu2022transfusion,gao2022multi} and investigate a large-scale study to validate our approach. We also plan to examine pathological identification based on the learned latent space.



\bibliographystyle{splncs04}

\bibliography{DeepRecon}

\end{document}